\documentclass[dvipsnames,sigconf=false, review=false, anonymous = false, nonacm=true]{acmart}

\usepackage{graphicx} 
\usepackage{subcaption}
\usepackage{xcolor}
\usepackage{algorithmicx}
\usepackage[noend]{algpseudocode}
\usepackage{algorithm}
\usepackage{xspace}
\usepackage{dsfont}
\usepackage{subcaption}
\usepackage{hyperref}

\usepackage{orcidlink}

\AtBeginEnvironment{quote}{\itshape}

\copyrightyear{2025}
\acmYear{2025}
\setcopyright{none}
\acmConference[FOGA XVIII]{2025 Foundations of Genetic Algorithms}{August 27--29,
2025}{Leiden, The Netherlands}
\acmBooktitle{2025 Foundations of Genetic Algorithms
 (FOGA XVIII), August 27--29, 2025, Leiden, The Netherlands}

\newcommand{\cma}{\textit{CMA-ES}\xspace}
\newcommand{\opocma}{\textit{(1+1)-CMA-ES}\xspace}
\newcommand{\kmean}{\textit{K-Means++}\xspace}

\newcommand{\vc}{\mathbf{c}}
\newcommand{\vx}{\mathbf{x}}
\newcommand{\vy}{\mathbf{y}}

\ccsdesc[500]{Theory of computation~Design and analysis of algorithms}
\ccsdesc[500]{Theory of computation~Bio-inspired optimization}

\begin{document}

\title{A Standardized Benchmark Set of Clustering Problem Instances for Comparing Black-Box Optimizers}
\author{Diederick Vermetten}
\affiliation{%
  \orcidlink{0000-0003-3040-7162}
  \institution{Sorbonne Université, CNRS, LIP6}
  \city{Paris}
  \country{France}
}
\email{Diederick.Vermetten@lip6.fr}

\author{Catalin-Viorel Dinu}
\affiliation{%
  \orcidlink{0009-0008-8490-9946}
  \institution{Sorbonne Université, CNRS, LIP6}
  \city{Paris}
  \country{France}
}
\email{Catalin-Viorel.Dinu@lip6.fr}

\author{Marcus Gallagher}
\affiliation{%
  \orcidlink{0000-0002-6694-9572}
  \institution{School of EECS, University of Queensland}
  \city{Brisbane}
  \country{Australia}
}
\email{marcusg@uq.edu.au}

\begin{abstract}
    One key challenge in optimization is the selection of a suitable set of benchmark problems. A common goal is to find functions which are representative of a class of real-world optimization problems in order to ensure findings on the benchmarks will translate to relevant problem domains. While some problem characteristics are well-covered by popular benchmarking suites, others are often overlooked. One example of such a problem characteristic is permutation invariance, where the search space consists of a set of symmetrical search regions. This type of problem occurs e.g. when a set of solutions has to be found, but the ordering within this set does not matter. 
    
    The data clustering problem, often seen in machine learning contexts, is a clear example of such an optimization landscape, and has thus been proposed as a base from which optimization benchmarks can be created. In addition to the symmetry aspect, these clustering problems also contain potential regions of neutrality, which can provide an additional challenge to optimization algorithms.
    
    In this paper, we present a standardized benchmark suite for the evaluation of continuous black-box optimization algorithms, based on data clustering problems. To gain insight into the diversity of the benchmark set, both internally and in comparison to existing suites, we perform a benchmarking study of a set of modular CMA-ES configurations, as well as an analysis using exploratory landscape analysis. Our benchmark set is open-source and integrated with the IOHprofiler benchmarking framework to encourage its use in future research.

\keywords{Benchmarking Aspects \and Continuous Optimization \and Fitness Landscapes \and Connection between BBO and ML} 
\end{abstract}
\maketitle

\section{Introduction}

Benchmarking is a key aspect in the development and analysis of optimization algorithms. Through rigorous benchmarking, we can gain insights into the relative strengths and weaknesses of different algorithmic ideas when applied to certain types of challenge, understand optimization behavior, and much more. Generally, we aim to benchmark either on a set of problems with well-understood structure, since this can lead to useful insight about algorithm behavior, or on problems which are assumed to match some aspect of specific real-world problem domains. 

It is not always straightforward to find benchmark problems which meet this criterion of representing specific types of challenges. While some aspects are well-covered by a variety of benchmarking suites, others might be much more sparsely distributed, and thus more difficult to analyze in detail. 

One such aspect that has been little studied is permutation invariance, which occurs for example when the `true' search should take place on a set of items, but is represented as an ordered list. This happens when the ordering between variables (or sets of variables) doesn't impact the evaluation function. In this case, there are natural symmetry regions in the space, which correspond to the same underlying solution space. 

Problems with permutation invariance exist in a wide variety of domains, ranging from facility location\cite{brimberg2008survey}, placement problems\cite{bliek2023benchmarking}, clustering\cite{gallagher2016towards} and neural network training\cite{hecht1989neurocomputing,chen1993geometry} to design optimization problems~\cite{otaki2022thermal}. In statistics and machine learning, this invariance is referred to as the label switching problem, for example in mixture models~\cite{pml1Book,stephens2000dealing}. 

There is a large amount of previous work where clustering problem instances have been used to evaluate the performance of a range of different types of global, black-box, metaheuristic and other optimization algorithms. Unfortunately, it is typically difficult to perform a meta-analysis or comparison of previous results across papers, because of issues such as different experimental settings, different performance metrics reported and different datasets utilized\cite{bartz2020benchmarking, lopez2021reproducibility}. Providing standardized sets of problem instances, making data and software available and integration with benchmarking tools is critical to support the research community in producing more comparable experimental results. In this paper, we introduce a benchmark suite based on these clustering problems which is integrated with the IOHprofiler framework\cite{wang2022iohanalyzer, de2024iohexperimenter} for usability. We also provide an interface to create custom clustering problems and analyze our proposed suite by performing several benchmarking studies on it and comparing it to COCO's well-known BBOB problem suite~\cite{hansen2021coco, bbobfunctions}. 

\section{The Clustering Problems}\label{sec:problems}

\begin{figure*}
    \centering
    \includegraphics[width=0.95\linewidth]{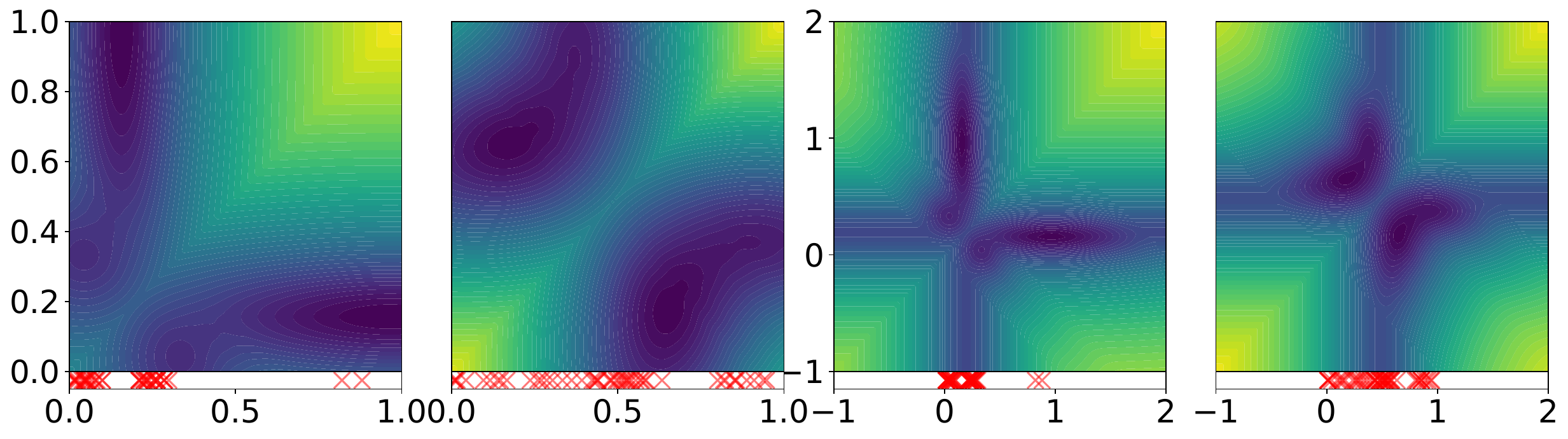}
    \caption{Contour plots of selected clustering problems based on 1-dimensional data (visualized as red crosses below the respective plots) with two cluster centers. The two right-most plots are based on the same data as the two left-most plots, but plotted using a wider domain.}
    \label{fig:1d_examples_wide}
\end{figure*}

To create our benchmarking suite, we make use of (centroid) data-clustering problems, which are commonly seen in machine learning and data analysis\cite{jain1999data}. We build on previous studies which showcase the potential use of these types of problems as optimization benchmarks~\cite{gallagher2016towards,rapin2019exploring}.
The Mean Squared Error (MSE) clustering problem can be defined as follows. Given a training set of $n$ data points, $ \mathcal{D} = \{\vx_1,\vx_2,...,\vx_n\} \subseteq \mathbb{R}^d $ , determine $k$ cluster centers $ C = \{\vc_1,\vc_2,...,\vc_k\} \in \mathbb{R}^d $ to minimize:

\begin{equation}
\begin{split}
f(C \mid X) = \frac{1}{n} \sum\limits_{i=1}^{n} \sum\limits_{j=1}^{k} b_{i,j} ||\vx_i - \vc_j ||^2
\end{split}
\label{sse}
\end{equation}
where 
\begin{equation}
 \begin{split}
b_{i,j} = 
\begin{cases}
    1   & \text{if }  ||\vx_i - \vc_j ||=min_j ||\vx_i - \vc_j || \\
    0   & \text{otherwise}
\end{cases}
\end{split}
\end{equation}

Note that the decision variables of the problem, $\vy$ are the $d$-dimensional coordinates of each cluster center, collected into a vector, meaning that the optimisation problem is of dimensionality $m \equiv kd$
\begin{equation}
    \vy = (y_1,\ldots,y_m) \equiv \vc = (c_1,\ldots,c_{kd}).
\end{equation}
The key parameters of this problem class are the dimensionality ($d$), cardinality ($n = |\mathcal{D}|$) and distribution of the data points and the number of cluster centers ($k$). 

For $k$ = 1, the fitness landscape is a (convex) quadratic bowl, but with higher values for $k$, the landscape becomes a combination of quadratic bowls of different sizes, intersecting with discontinuous ridges\cite{gallagher2016towards}. Consequently, the landscapes of clustering problems are in general non-linear and non-convex with a large number of local optima.

Using MSE implies that the (squared) Euclidean distance measure is used. Since the choice of distance and error measure could potentially have an impact on the resulting optimization landscape, we provide a flexible problem generator where these design choices can be modified in the future. However, we stick with these default values for the provided benchmark suite and for any experiments presented in this paper.

Given this definition, we can create instances of clustering problem by defining datasets $\mathcal{D}$ and a corresponding number of cluster centers to find $k$. In Figure~\ref{fig:1d_examples_wide}, we show the contour plots of fitness landscapes for two datasets with $d=1$ and $k=2$.

\subsection{Symmetries}

A key property of using clustering problems as optimization benchmarks is that these problems have inherent symmetries that are not typically found in other (continuous) benchmark problem sets. The ordering of cluster centers in $C$ when concatenated into $\vy$ is arbitrary. In general, this means that for any clustering problem, there are $k!$ symmetric regions in the search space. This type of symmetry is also found in some combinatorial problems such as graph coloring~\cite{tayarani2013landscape}. In Figure~\ref{fig:1d_examples_wide}, the symmetry can be seen as a diagonal ridge dividing the search space in two halves since $k=2$.

The question of symmetry is present in a wide variety of optimization problems, as discussed earlier, and there has thus been some interest in modifying optimization algorithms to handle this specific challenge. In Bayesian optimization, recent studies have proposed modifications of the kernel function to account for the permutation-symmetries present in these search spaces, resulting in promising improvements in performance~\cite{brown2024sample,fotias2024optimization}. In the evolutionary algorithms community, questions of invariance in the search space have focused on transformations such as rotation and translation~\cite{hansen2008adaptive}, but the question of permutation invariance has not been studied as broadly. 

\subsection{Neutrality}

Another important property of clustering problems is that the fitness landscapes contain a significant amount of neutrality. Neutrality has been studied in fitness landscape analysis and across a number of problem domains, predominantly in the discrete case (see, e.g.~\cite{verel2010local} and the references therein). If an optimization problem is formulated with one or more decision variables that have no/negligible effect on the objective function, this will create a search space with neutrality. Therefore, it seems reasonable to expect that this is a property of some continuous real-world optimization problems. 

As formulated above, the clustering problem is unconstrained, but the range of each data variable provides a soft boundary constraint that can be used to initialize an algorithm, since a solution will not be improved by moving cluster center outside of the range of the data. If one (or more) cluster centers, $\vc_j$ is not the closest cluster center for any of the data points $\vx_i$ (i.e. for $\vc_j$, $b_{i,j} = 0 \; \forall\; i \in \{1,\ldots,n\}$) it means that the landscape contains regions that are perfectly flat. This can be seen in the two right-most plots of Figure~\ref{fig:1d_examples_wide}, where the search ranges have been extended beyond the original data points, revealing regions of neutrality.

Note that additional conditions can be added, such as requiring each cluster center to be closest to at least one data point and that each data point is closest to only one cluster center (i.e. non non-equidistant), see, e.g.\cite{bagirov2008modified}. However, this would require constraint handling mechanisms to be incorporated into the optimization algorithms, therefor these constraints are not generally included in the formulation of these clustering optimization problems.

\begin{figure*}[t]
    \centering
    \includegraphics[width=0.95\textwidth]{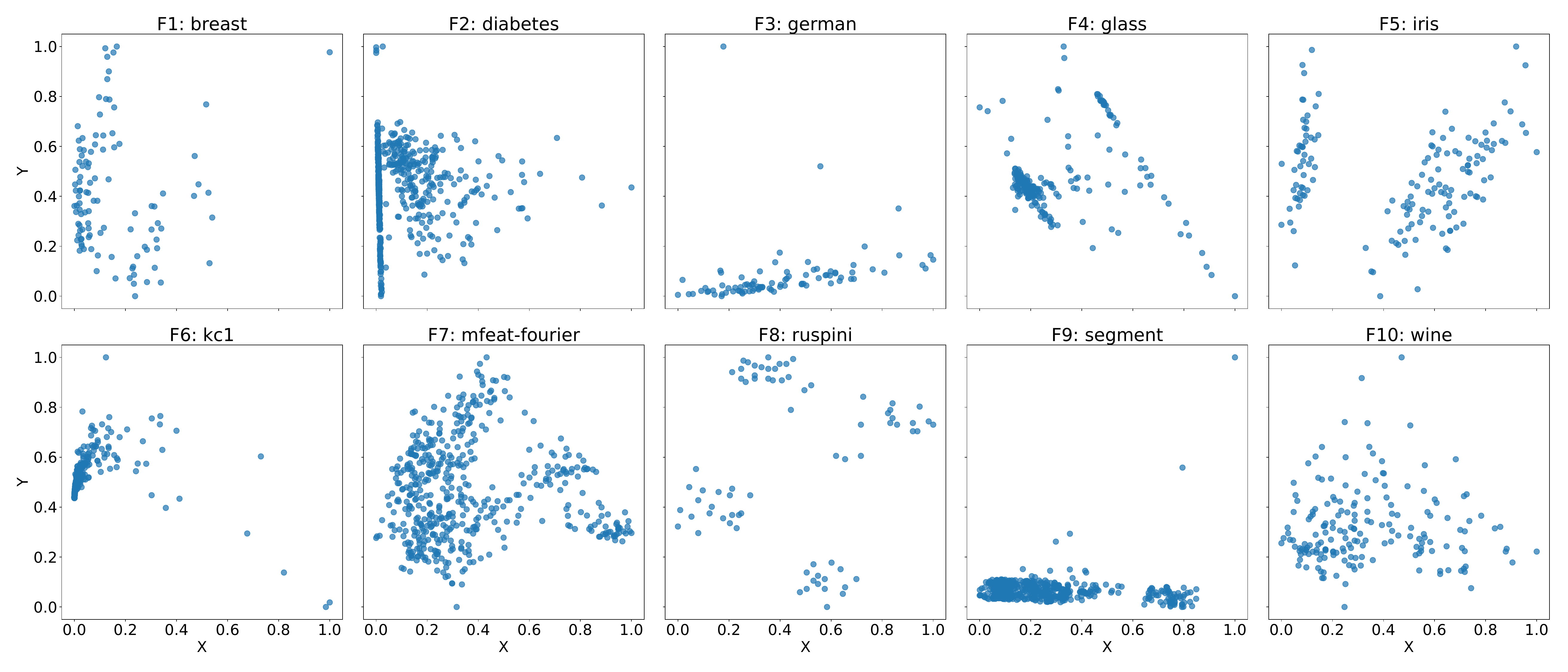}
    \caption{Data underlying each of the 10 problem sets.} 
    \label{fig:datasets}
\end{figure*}

\subsection{Benchmark Suite}
Clustering is a widely-studied problem, and a wide variety of different datasets have been utilized to evaluate the performance of clustering algorithms (e.g.\cite{ClusteringDatasets}). Much of this work is not focused on evaluating the quality of the solutions found in terms of the optimization objective function, but rather uses metrics to evaluate the quality of the data clustering produced, as well as the performance of using the clustering as a classifier (assuming the data contains labels for true classes). Many papers do not report the objective function values found, limiting the ability to assess the performance of the underlying optimizer.

A set of 27 problem instances was produced and used in~\cite{gallagher2016towards}, with Matlab code and information such as (approximate) global optimum solution vectors and objective values available at: \url{https://marcusgal.github.io/ess_clustering.html}. However, one limitation is that the problem instances vary over many dimensionalities, while it is often useful to compare optimizers on a set of problems of a fixed dimension. Two of these problem instances were also included in the MLDA~\cite{kerschkemachine,rapin2019exploring}, incorporated into Nevergrad~\cite{bennet2021nevergrad}.

To create our benchmark suite, we selected a total of 10 commonly-used datasets from different machine learning contexts. To enable consistent problem dimensionalities for e.g. aggregation of performance data, we opted to reduce all data-spaces to 2 dimensions using principal component analysis (PCA). Since we opt to use $k\in\{2,3,5,10\}$, this leads to problem dimensionalities of $n\in\{4,6,10,20\}$ for each of the 10 datasets.

To provide consistent search domains for the optimization algorithms, we min-max normalize the data-space (after PCA) to $[0,1]^n$. The resulting datasets are visualized in Figure~\ref{fig:datasets}. These figures show that there is quite some variety between the datasets in terms of the density of datapoints and the presence of outliers. 

\subsection{IOHClustering Python Package}

To ensure the accessibility of our proposed benchmark suite, we created a new Python package 'IOHclustering', which integrates with the existing IOHprofiler framework. This allows users to rely on the logging methodology from IOHexperimenter~\cite{de2024iohexperimenter}, and subsequently analyze and visualize their performance data using tools such as IOHanalyzer~\cite{wang2022iohanalyzer} and IOHinspector~\cite{vermetten2025mo}. 

In addition to the suite of problems described above, IOHclustering also integrates a generic way to instantiate arbitrary clustering problems by providing a dataset and a number of cluster centers. The resulting problem can be further customized by changing the distance metric used for cluster assignment and/or the error metric used to judge the cluster quality. As such, new instances can be generated from arbitrary combinations of these four parameters.

\section{Analysis of Benchmark Suite}

When introducing a new problem suite, it is important to consider its usefulness for benchmarking tasks. To this end, we consider the desirable properties for benchmarking problem sets as described in~\cite{bartz2020benchmarking}, which are as follows:
\begin{itemize}
    \item Diversity: the problems in the suite should be sufficiently varied, ideally from both an algorithm performance perspective (easy / hard) and from a more fundamental problem characteristics perspective (different kinds of optimization challenges). 
    \item Representativeness: the problems in the suite should contain challenges which are likely present in a set of real-world optimization problems.
    \item Scalability and Tunability: Ideally, a problem suite / framework should make it possible to tune the characteristics of the problems, whether on a high level (e.g. number of variables) or on a lower level (e.g. variable interactions, degree of multimodality)
    \item Known solutions: If the global optima of the problems are known, it makes it easier to compare performance of the algorithms. If no exact values are known, strong best-known results could be useful alternatives. 
\end{itemize}

When considering these criteria, we note that the question of representativeness is the driving motivation for this suite. As discussed in Section~\ref{sec:problems}, the inherent symmetry is a problem characteristic which occurs in a variety of domains, but is not widely covered by optimization benchmarks. 
\begin{figure*}[t]
    \centering
    \begin{subfigure}[t]{0.3\linewidth}
        \centering
        \includegraphics[width=0.95\linewidth]{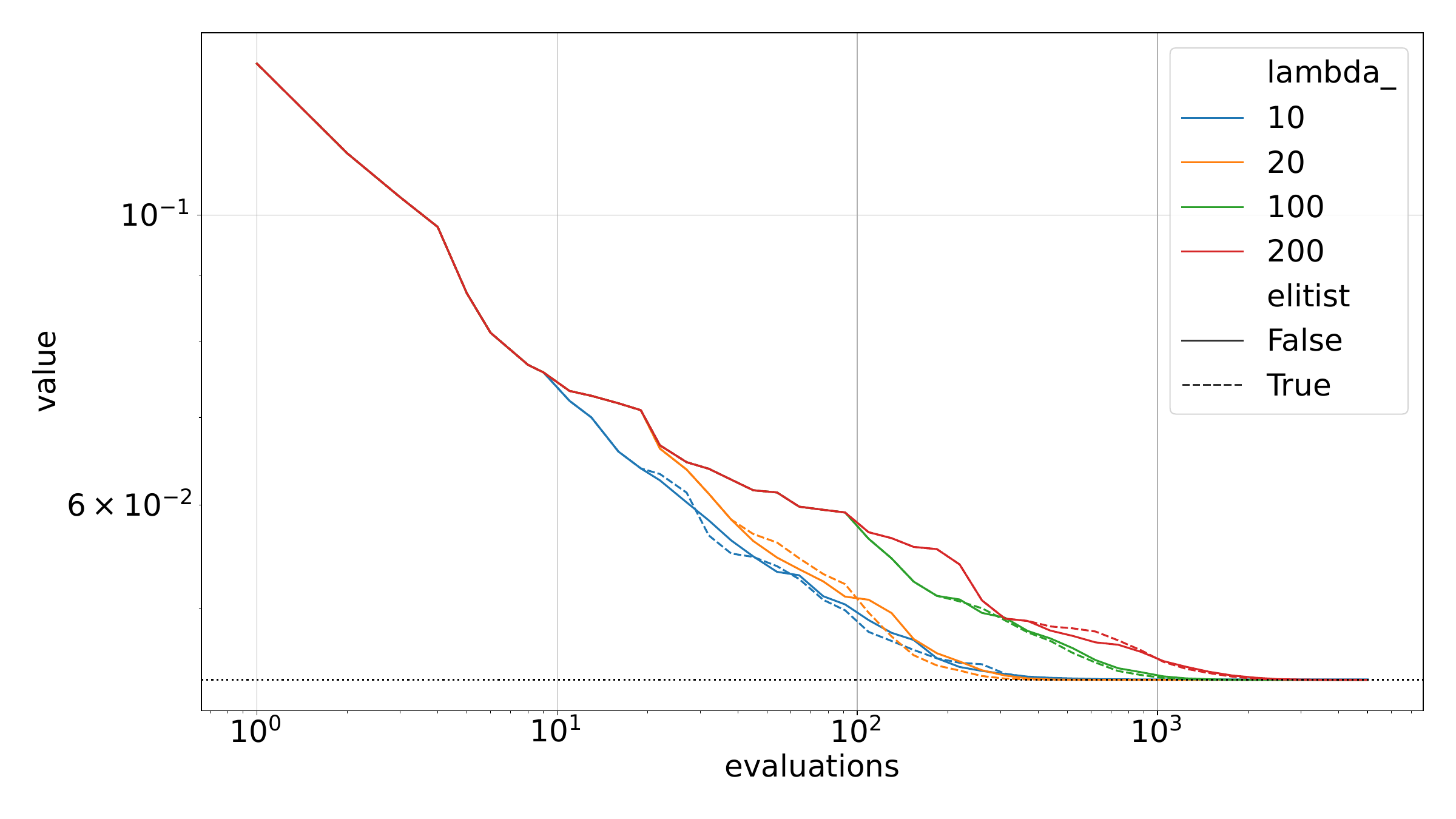}
        \caption{Function 1, k=2}
    \end{subfigure}
    \begin{subfigure}[t]{0.3\linewidth}
        \centering
        \includegraphics[width=0.95\linewidth]{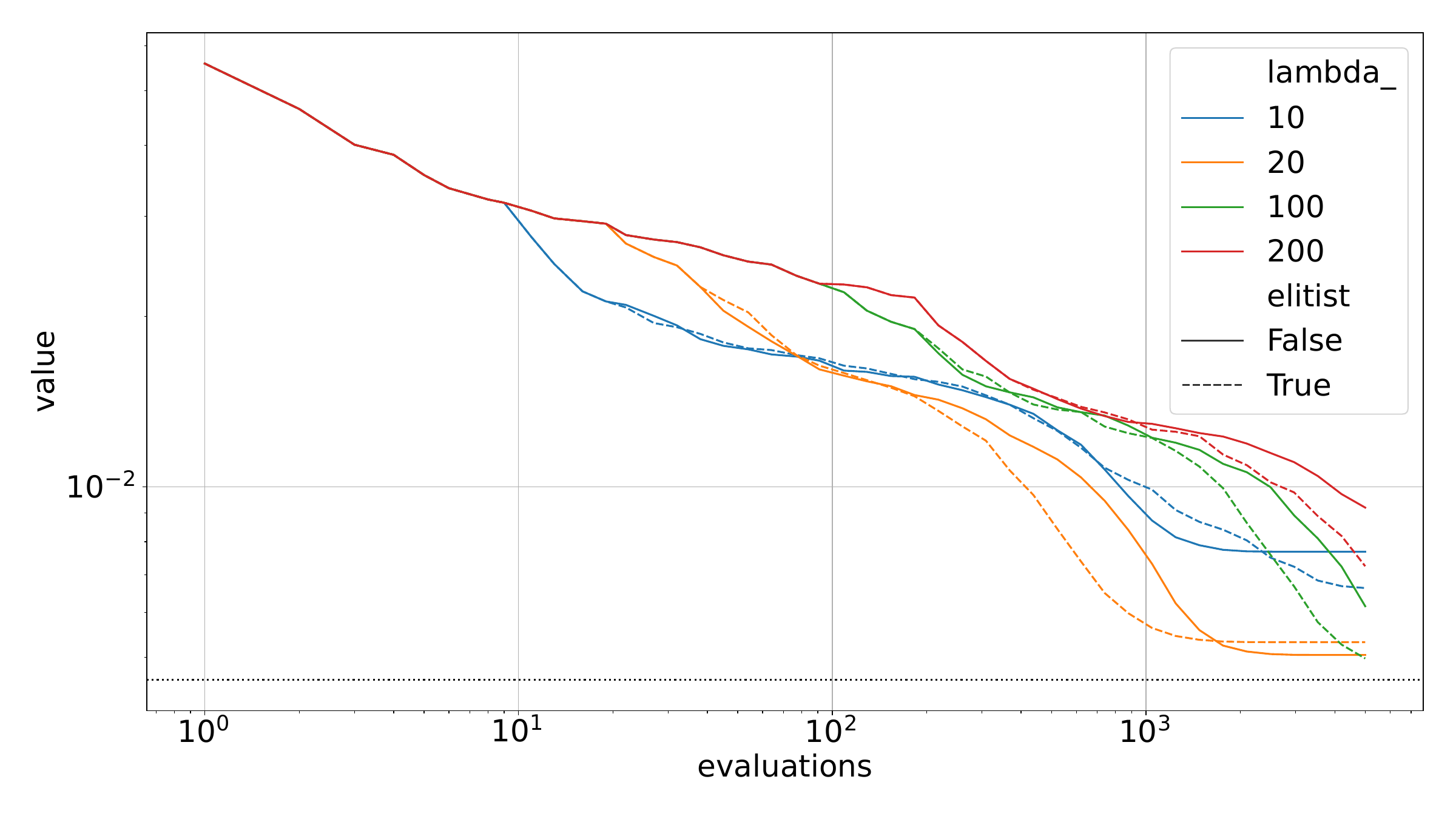}
        \caption{Function 1, k=10}
    \end{subfigure}
    \begin{subfigure}[t]{0.3\linewidth}
        \centering
        \includegraphics[width=0.95\linewidth]{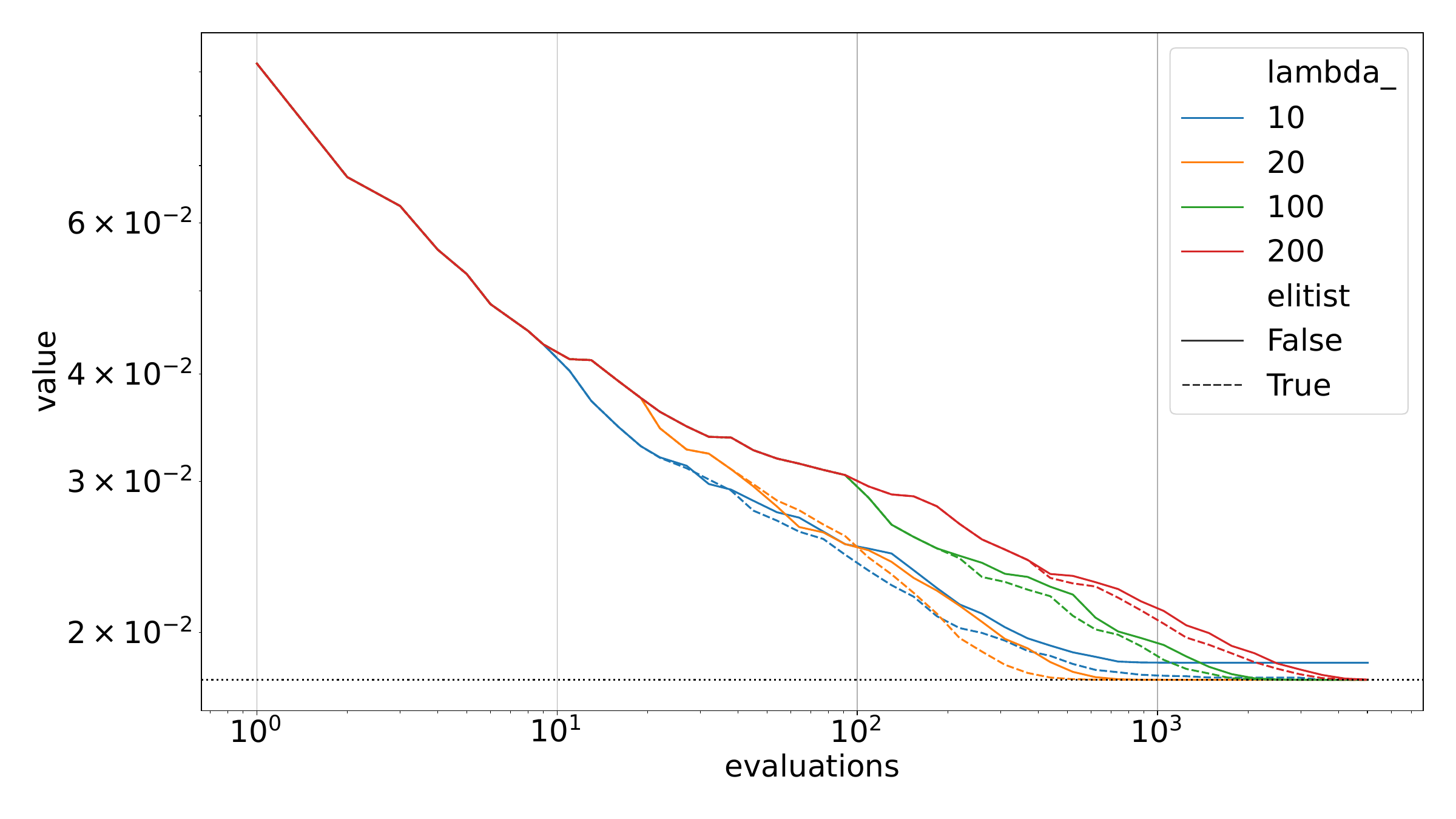}
        \caption{Function 2, k=3}
    \end{subfigure}
    \caption{Some example convergence curves (geometric mean) for different population sizes and elitism options (bound correction off, covariance on, $\mu=10$). The dotted line represents the best value found by 100 repetitions of \kmean. }
    \label{fig:conv_curves}
\end{figure*}

The question of scalability is incorporated into the problem suite in the form of the number of clusters $k$, which directly impacts the number of variables of the optimization problem. Note that this is a different type of scalability as commonly found in e.g. BBOB, since we don't have any guarantees that the optimization landscape for a given dataset and $k$ value is similar to the landscape for the same dataset with changed $k$. 

Regarding tunability, we should note that a benchmark suite of fixed size will necessarily be limited in this regard. However, through providing an accompanying problem generator, we hope to facilitate the creation of clustering problems with more flexible problem characteristics. Some work has considered the idea of searching the problem space by modifying points (or properties of their distribution) to create problems that are more challenging, or provide differentiation between different algorithms\cite{gallagher2019fitness,hajari2024searching}.

While our problem suite does not include known optima, we can find very strong baseline values for solution quality. Since we are dealing with clustering problems, we can get these performance baselines by running the well-known (non-black-box) \textit{$k$-Means} algorithm, more specifically with \kmean initialization as used in many software packages by default\cite{arthur2006k}. Since this algorithm is stochastic, we perform 100 repetitions on each problem in our suite to collect baselines. For ease-of-use, these values are also accessible in our Python package.

\subsection{Benchmarking Study}\label{sec:benchmarking}

While the previous criteria for benchmark suite quality were relatively straightforward to address, the question of problem diversity requires a more involved experimental setup. We can judge diversity in different ways: diversity within the benchmark suite itself, and diversity with respect to existing benchmark suites. Subsequently, this diversity can be further split into diversity of algorithm performance and diversity of problem characteristics. We will address each of these aspects, starting with algorithm performance within the suite. 

To analyse algorithm performance, we run a benchmark study using a large set of configurations of the \cma algorithm~\cite{hansen2001completely}. We make use of the Modular \cma framework~\cite{de2021tuning} for this purpose, and create a set of 128 configurations by modifying the following parameters:
\begin{itemize}
    \item Covariance adaptation: On / Off
    \item Elitism: On/ Off
    \item Boundary Correction: Off / Saturate
    \item Lambda: 5, 10, 20, 100, 200
    \item Mu: 5, 10, 20, 50, 100
\end{itemize}
Note that while the full enumeration of these settings results in a slightly larger set of configurations, we need to have $\lambda\geq\mu$ in order to have a valid algorithm. 
Each of the 128 configurations is run on each of the 40 benchmark problems, with 25 independent repetitions of budget $5000$ (not scaled by problem dimensionality). We fixed the random seeds to ensure all configurations with a given population size start from the same set of samples. 

To ensure reproducibility of this benchmarking setup, and all other experiments discussed in this paper, we provide a Zenodo repository which contains the script, data and visualization methods required to reproduce all presented results, as well as several additional figures~\cite{zenodo}.

\begin{figure}
    \centering
    \includegraphics[width=0.95\linewidth]{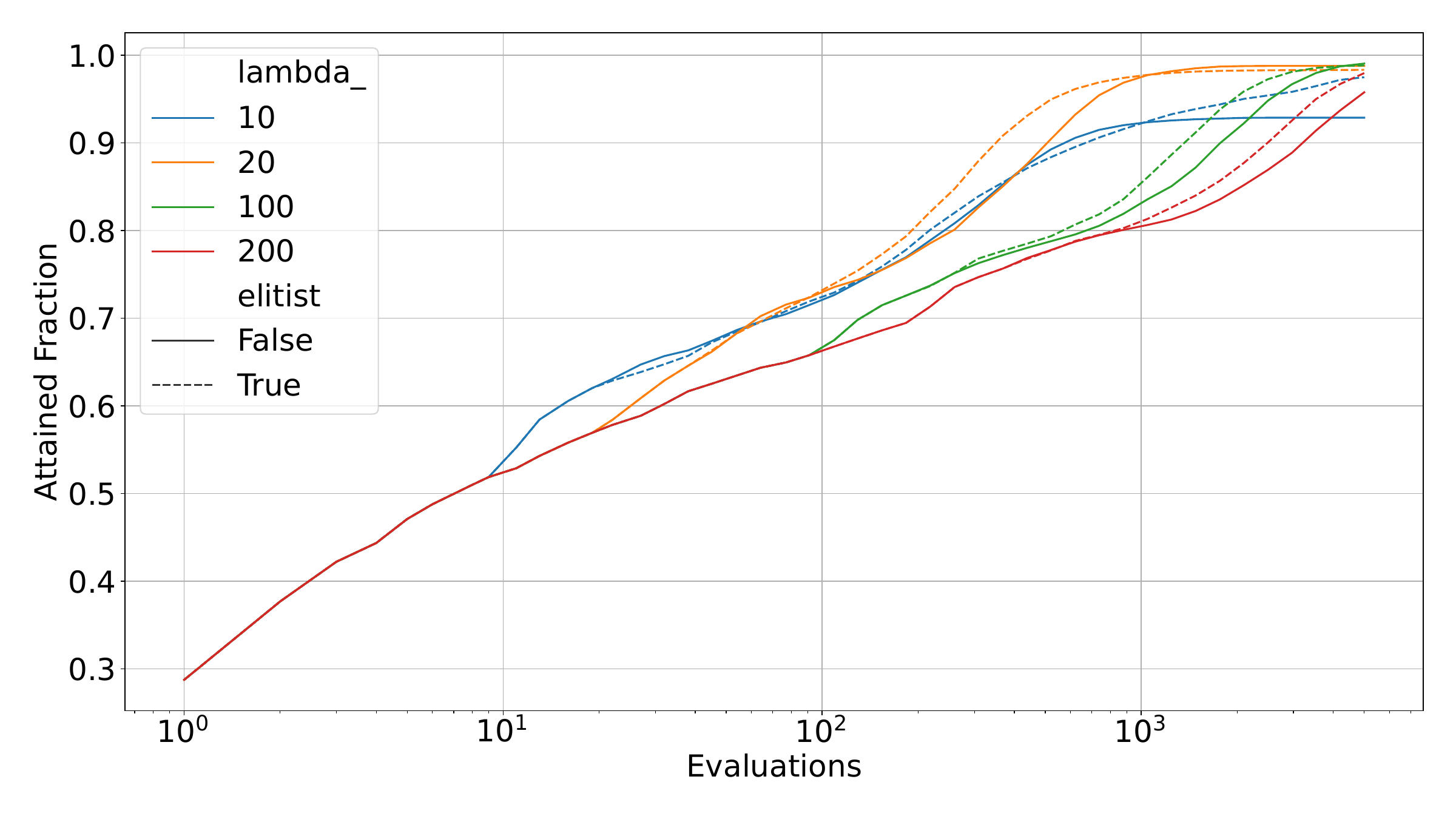}
    \caption{EAF over all 10-dimensional problems in our suite, for different population sizes and elitism options (bound correction off, covariance on, $\mu=10$). The EAF bounds are set to the \kmean baseline and the worst seen value by all \cma variants, respectively (with a log-scaling between them). }
    \label{fig:eaf_10d}
\end{figure}

First, we analyze the convergence behavior of a smaller set of \cma configurations. Figure~\ref{fig:conv_curves} shows the convergence curves for 3 different benchmark problems, where for each problem, 8 configurations are selected by varying $\lambda$ and elitism. The equivalent convergence curves for all other problems are available in our Zenodo repository~\cite{zenodo}. From these convergence curves, we observe that for low-dimensional problems, the total budget provided is sufficient for all selected \cma configurations to converge to the same performance value. However, differences between configurations are clearly present in terms of convergence speed, with the lower population sizes resulting in notably faster convergence. However, as dimensionality increases, differences between configurations become more pronounced and lower values of $\lambda$ become more prone to stagnation.

In addition to the convergence curves, the fact that we have baseline performance values from \kmean means that we can create aggregations over different datasets. We can do this in the form of the Empirical Attainment Function (EAF), with its bounds chosen to be the \kmean value and the worst-observed value in all \cma runs respectively. Note that this corresponds to the ECDF between these bounds if we assume an infinite number of targets~\cite{lopez2024using}. The EAF aggregated over all Datasets with $k=5$ is shown in Figure~\ref{fig:eaf_10d}. 

One clear result from analyzing the EAF in Figure~\ref{fig:eaf_10d} (and the convergence curves in Figure~\ref{fig:conv_curves}) is the impact of the overall budget on the relative ranking of different configurations. While the remaining analysis of algorithm performance focuses on the fixed-budget perspective (final performance after $5000$ evaluations), this serves as a proof-of-concept to illustrate the proposed benchmark problems, not as a judgment for which algorithm configuration is the most appropriate for this suite.

\begin{figure}
    \centering
    \includegraphics[width=\linewidth]{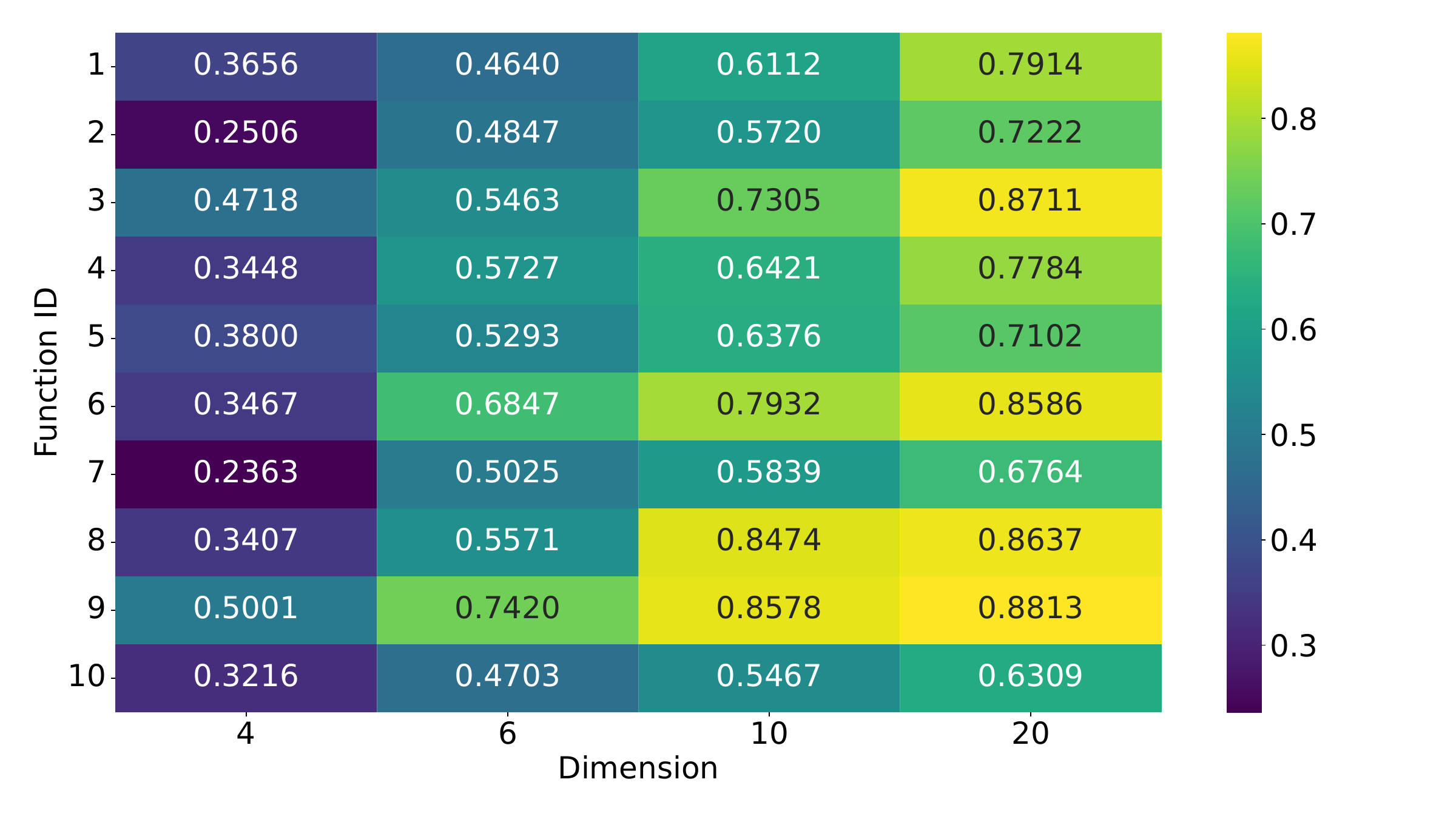}
    \caption{Relative differences (in terms of function average final function value) between best and worst \cma configurations in the portfolio.}
    \label{fig:reldifffs}
\end{figure}

To better understand whether our benchmark suite is able to distinguish between algorithms, we zoom in on the overall differences in performance between all algorithms in our portfolio. In particular, we compare the performance of the best and worst algorithm configurations to gauge the breadth of performance found on each problem. In Figure~\ref{fig:reldifffs} we show these differences for each problem in our proposed suite. From this figure, we can see that the main factor which determines the scale of performance differences is the problem dimensionality. However, even for the setting with dimensionality $4$, there are still reasonable differences present between the best and worst versions of the \cma. When comparing differences between functions of the same dimensionality, we observe there are also some visible patterns, with the relative ordering of the performance differences remaining relatively stable across dimensionalities.

\begin{figure}
    \centering
    \includegraphics[width=0.95\linewidth]{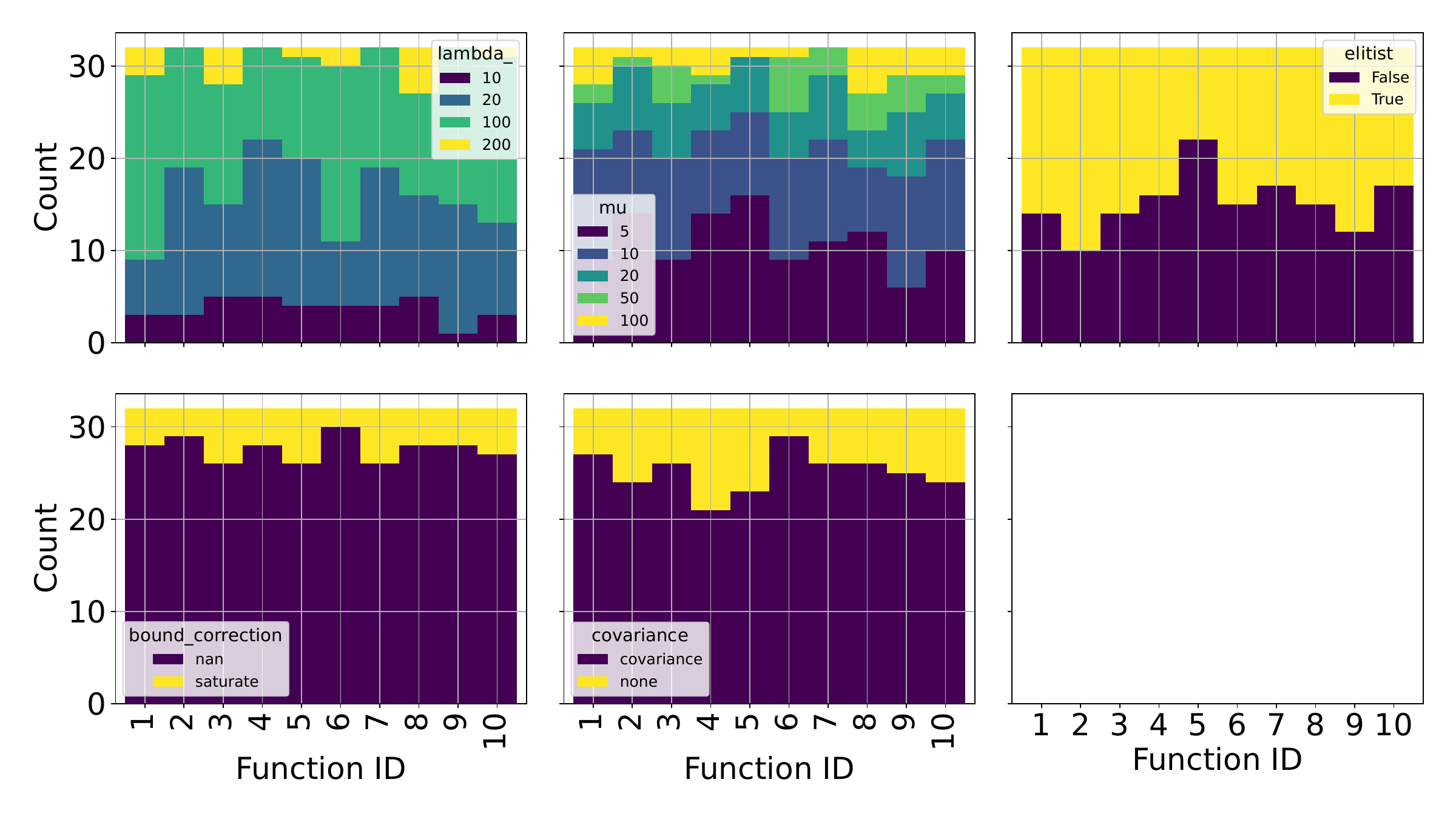}
    \caption{Distribution of parameters in the top 8 (out of 128) configurations on each function, grouped by dataset. }
    \label{fig:rank_modules}
\end{figure}

Next, we look for patterns in the top-ranking configurations on each benchmark problem. To this end, we create a stacked bar chart of module counts in the top 8 configurations on each problem. In Figure~\ref{fig:rank_modules} we show these results aggregated by dataset. From this, we see that the overall differences between datasets are relatively small, with similar distributions of best-performing parameter settings. While the choice of elitism seems to be rather evenly split between top performing configurations, both the boundary correction and covariance modules have a setting with is much more common than its alternative.

For boundary correction, most top-performing configurations avoid using the saturation mechanism; instead, they ignore the boundaries entirely. This suggests that the regions outside of the box spanned by the datapoints might still contain useful information for the search process. For the covariance parameter, it is most commonly enabled, indicating that there might be some interactions between variables for which adapting the covariance matrix is beneficial.

\begin{figure}
    \centering
    \includegraphics[width=0.95\linewidth]{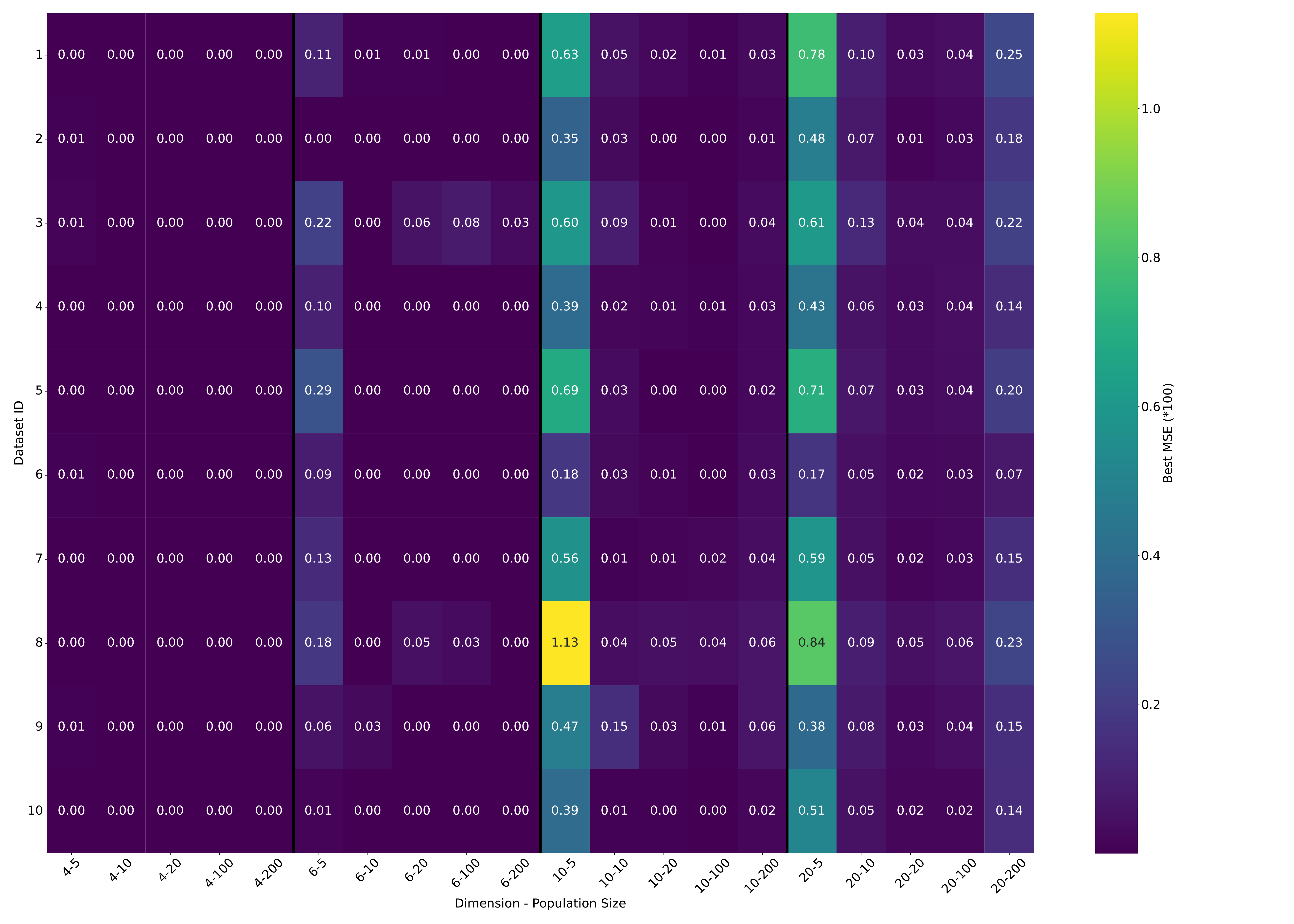}
    \caption{For each problem, the performance (average final function value) found by the best \cma configuration with the specified $\lambda$ value, shown as difference to the minimal value found by 100 repetitions of \kmean. }
    \label{fig:lambda_fval_rel}
\end{figure}

\begin{figure*}
    \centering
    \begin{subfigure}[t]{0.49\linewidth}
        \centering
        \includegraphics[width=0.99\linewidth]{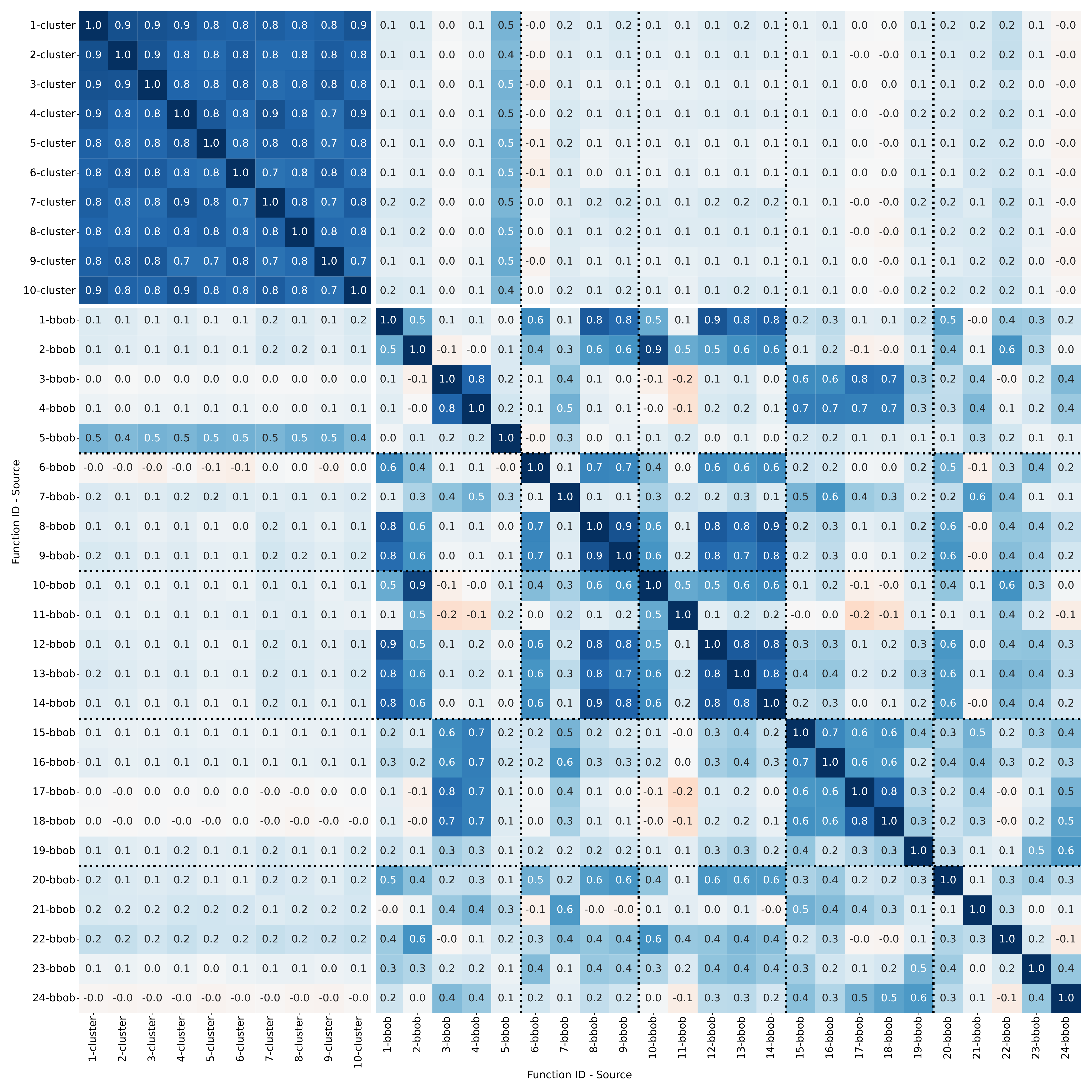}
        \caption{Kendall's $\tau$ correlation based on performance \\(best-so-far) of all 128 modCMA configurations. }
        \label{fig:correlation_perf}
    \end{subfigure}
    \begin{subfigure}[t]{0.49\linewidth}
        \centering
        \includegraphics[width=0.99\linewidth]{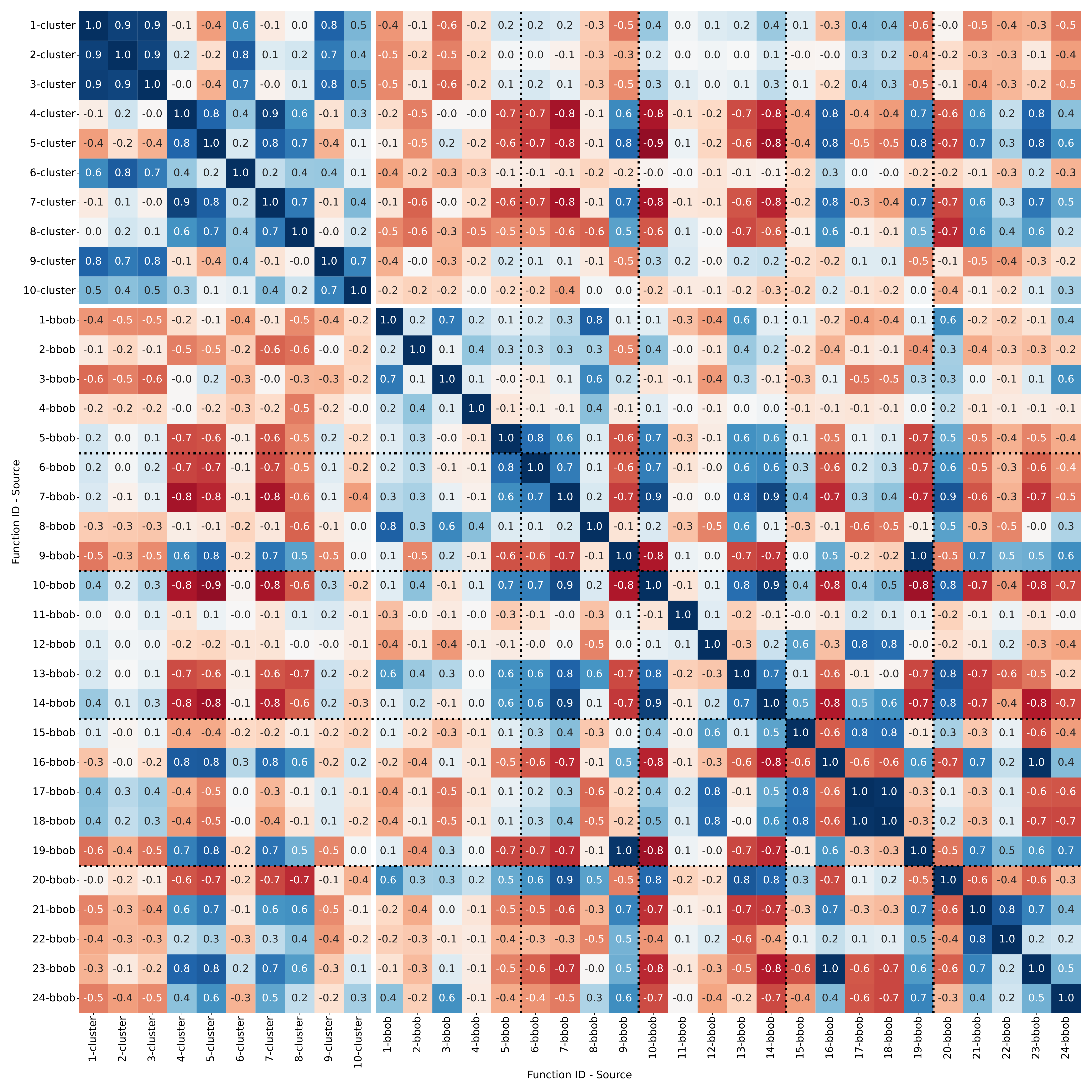}
        \caption{
        Cosine similarity between (here: standardized to 0 mean, variance 1; [0,1] normalized version available as well) ELA feature vectors (linmodel, meta, distr, level, pca, nbc and ic sets).}
        \label{fig:correlation_ela}
    \end{subfigure}
    \caption{Pairwise similarities between the clustering and BBOB problems. Shown here for dimensionality 10, other dimensionalities are available in our Zenodo repository~\cite{zenodo}.}
    \label{fig:correlation}
\end{figure*}
Finally, we compare the performance of our optimization algorithms to the baseline values found by \kmean. Here, we focus on the impact of population size, by taking the best-performing configuration for each population size and showing the absolute deviation to the K-means result. 
These differences are shown in Figure~\ref{fig:lambda_fval_rel}, where we notice that \cma reaches values which are very close to those found by K-means. This seems to confirm that our earlier observations on consistent convergence in low dimensionality corresponds to the (likely) global optimal solution. This seems to indicate that the CMA-ES can serve as a rather effective baseline algorithm for future comparison of optimization algorithms on this benchmark suite, while still showing potential for further refinement.

\subsection{Comparison to BBOB}

In order to understand whether our proposed benchmark suite does indeed result in different algorithmic challenges than existing benchmarking suites, we compare our problems to the equivalent-dimensional problems from the BBOB suite. We identify two main axes along which we can assess problem similarity: algorithm performance and landscape characteristics. From the algorithm performance perspective, we utilize the relative performance rankings of the CMA-ES configurations from Section~\ref{sec:benchmarking}, and compare this to their equivalent ranking on COCO's BBOB suite~\cite{hansen2021coco} (where we thus use problem dimensionalities $\{4,6,10,20\}$). While the BBOB suite has an instance generation mechanism~\cite{bbobfunctions}, this is not present in our suite, and thus we stick with only the first instance as a representative of each BBOB function.

For each problem, we create the ranking of the 128 \cma configurations based on the best average final fitness value found. Based on these rankings, we calculate Kendall's $\tau$ correlation~\cite{kendall1948rank} for each pair of problems, and visualize these correlations in Figure~\ref{fig:correlation_perf}. In this figure, the clustering problems are clearly separated from the BBOB functions, with high correlation of the CMA-ES configuration rankings between all 10 clustering problems. However, the correlations to the BBOB problems are generally much lower, with a peak for $F_5$, the linear slope. This could point to two underlying factors: on the one hand, more exploitative versions of \cma tend to work well on the linear slope, and based on the results presented in Figure~\ref{fig:lambda_fval_rel}, the same seems to be true for the clustering problems. However, since we don't observe the same high correlation to $F_1$, it might also be related to the fact that the boundary correction method is generally disadvantageous for the linear slope, which, according to Figure~\ref{fig:rank_modules}, is also the case on the clustering problems. 

Overall, Figure~\ref{fig:correlation_perf} shows that the clustering problems have high internal problem similarity, but differ from most of the problems present in BBOB. However, similarity in algorithm performance does not necessarily imply similarity in problem structure. To further asses the landscape differences between these problem sets, we can employ exploratory landscape analysis (ELA)~\cite{mersmann2011exploratory}. To collect ELA features, we generate a set of $4096$ samples 
in the search space (using a Sobol sampler). We then min-max normalize the function values to $[0,1]$ as recommended in \cite{prager2023nullifying}, use PFlacco~\cite{prager2024pflacco} to calculate the ELA features from the commonly used feature sets (linear model, meta, distribution, PCA, nearest best clustering, information content; as used in e.g. \cite{long2022learning}) 
and normalize the resulting feature values (mean 0, standard deviation 1). Notably, we don't perform feature selection (except for the standard removal of feature-set calculation times). 

Based on the normalized ELA features, we use cosine similarity on each pair of problems. The resulting heatmap is shown in Figure~\ref{fig:correlation_ela}. From this figure, we notice that the similarities between the set of clustering problems are still generally positive, but show more diversity. The similarity between clustering and BBOB problems gives more evidence for the claim that these problem sets are quite different, with most values moderately negative. There are outlying values for some of the more multi-modal problems, which could be expected in this problem dimensionality, as we have at least $5!=120$ equivalent global minima (since we have $k=5$ cluster centers). 
When considering other problem dimensionalities (and thus differing numbers of symmetries), as we make available in our Zenodo repository\cite{zenodo}, we observe that this observation becomes stronger for larger $k$, while the lower $k$-values show less clear patterns in terms of which BBOB-functions are more similar to our clustering problems. 

\subsection{Analysis of Optimization Landscapes}

To gain a deeper understanding of the landscapes induced by our clustering problems, we investigate the property of multi-modality. While the presence of multi-modality arising from problem symmetries is immediately apparent, it remains unclear whether each symmetry region contains only one or multiple local minima. To address this question, we use several local optimization algorithms, each initialized from 50 distinct points within the search space. These initial points are chosen such that they lie within the same symmetry region (enforced by ordering the first coordinate of the cluster centers).

We consider the following local search methods:
\begin{itemize}
    \item Powell’s method (\cite{powell1964efficient}, implementation from scipy~\cite{virtanen2020scipy}),
    \item L-BFGS-B (\cite{zhu1997algorithm}, implementation from scipy~\cite{virtanen2020scipy}),
    \item \opocma (discussed before, implementation from~\cite{de2021tuning}), with an initial step size $\sigma_0=0.1$ to promote local search behavior.
\end{itemize}

\begin{figure}
    \centering
    \includegraphics[width=\linewidth]{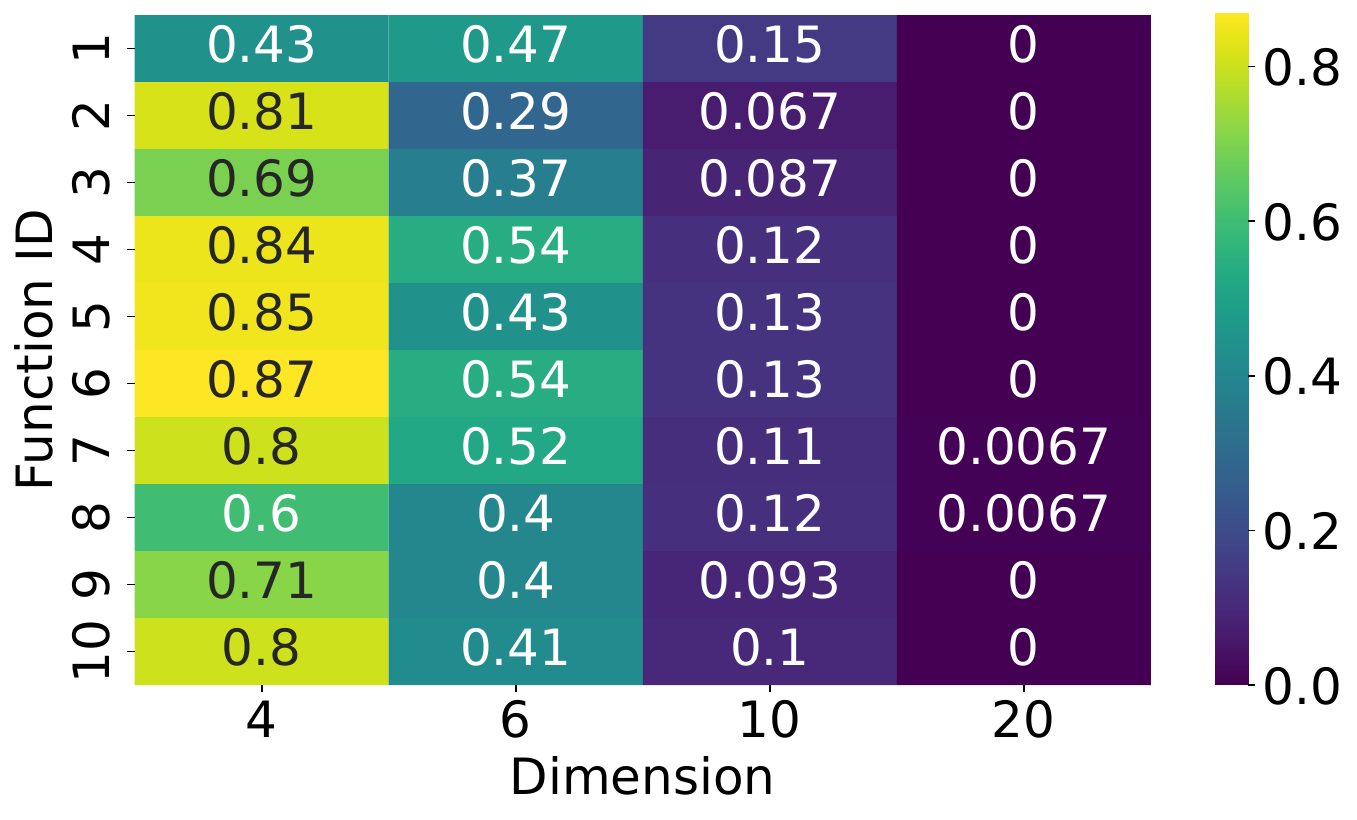}
    \caption{Fraction of solutions that ended up in the same region of symmetry as the starting point of the local search.}
    \label{fig:solution_percentage}
\end{figure}

In addition to probing the multi-modal structure, this setup allows us to examine the extent to which local search methods remain confined within their initial symmetry regions. Figure \ref{fig:solution_percentage} reports the proportion of solutions that remained within their original regions throughout the optimization process. As the dimensionality of the problem increases, the diversity of the final solutions also increases. In low-dimensional settings, such as for $k=2$, approximately 75\% of the solutions remain within the same symmetry region as their initial starting points after local search. However, as the dimensionality grows, this proportion decreases significantly. At 10 dimensions, only around 11\% of the solutions remain within their initial region, and at 20 dimensions, virtually no solutions are found to stay within their original symmetry regions. 

\begin{figure*}
    \centering
    \begin{subfigure}[t]{0.39\linewidth}
        \centering
        \includegraphics[width=0.9\linewidth]{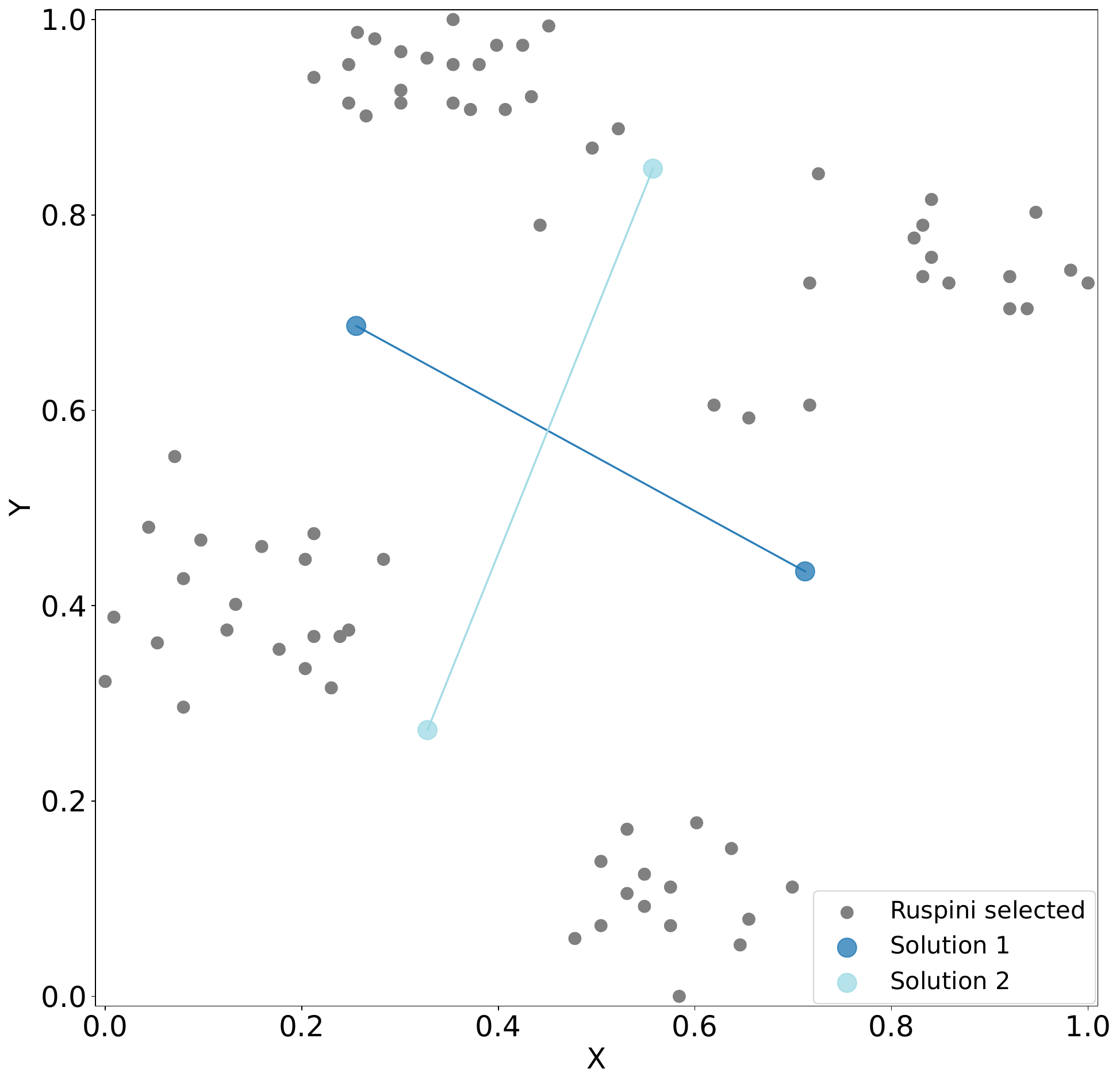}
        \caption{Distinct solutions found by local search \\ algorithms on function F8 with k=2 }
        \label{fig:ruspini-k2}
    \end{subfigure}
    \begin{subfigure}[t]{0.39\linewidth}
        \centering
        \includegraphics[width=0.9\linewidth]{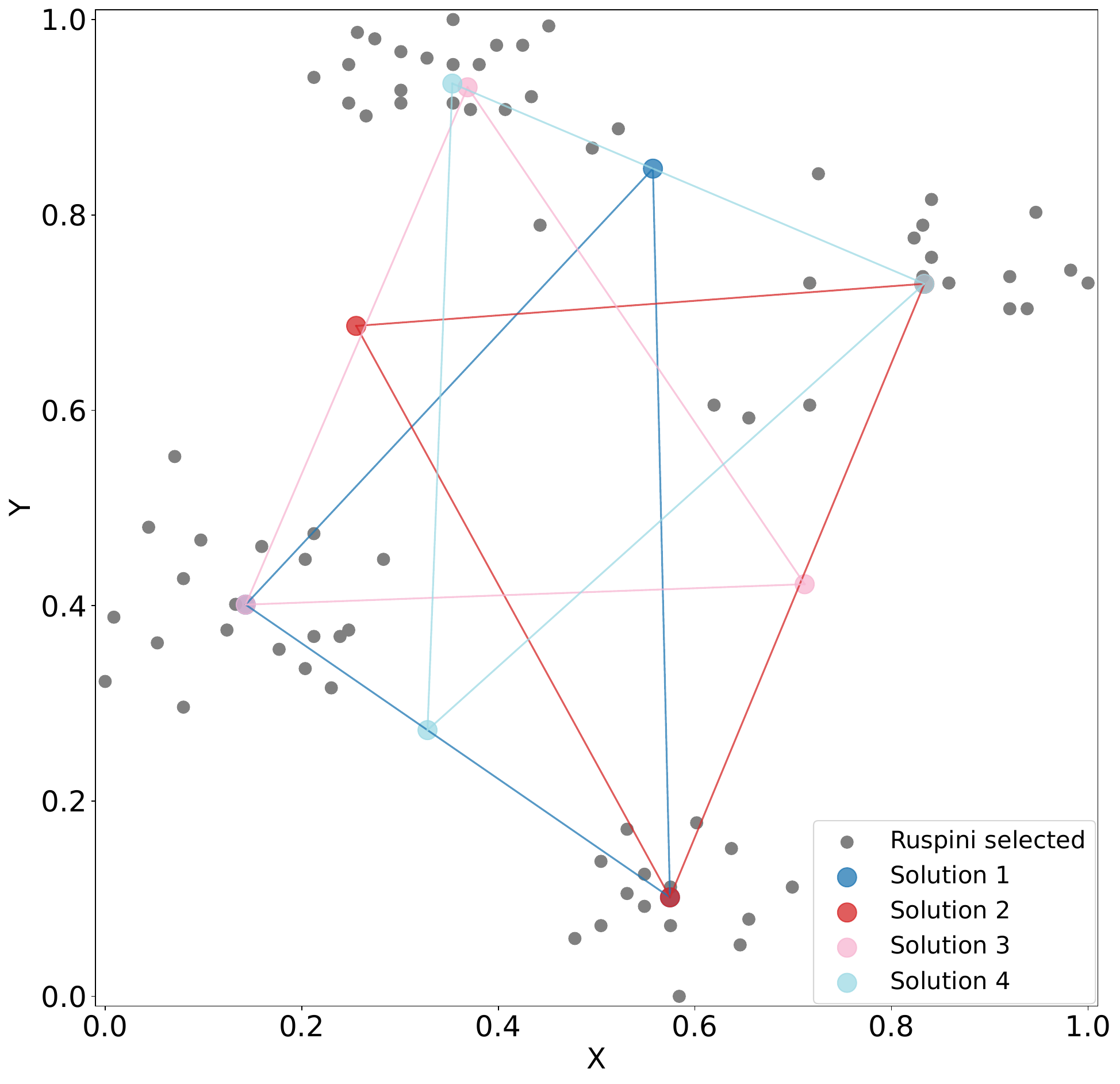}
        \caption{Distinct solutions found by local search \\ algorithms on function F8 with k=3 }
        \label{fig:ruspini-k3}
    \end{subfigure}
    \begin{subfigure}[t]{0.2\linewidth}
        \centering
        \includegraphics[width=0.8\linewidth]{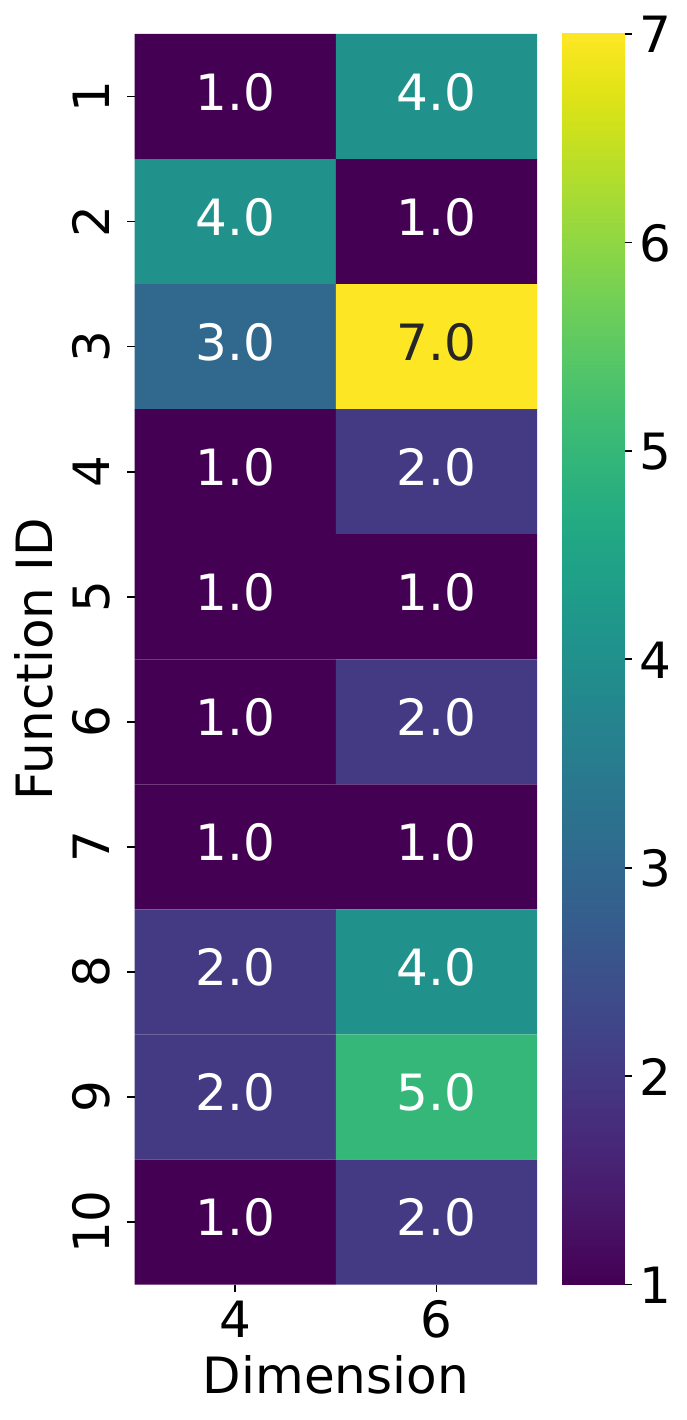}
        \caption{The number of distinct solutions found by local search algorithms}
        \label{fig:heatmap-diversity}
    \end{subfigure}
    \caption{Number of distinct local minima discovered by local search algorithms across different problem instances. The examples for $k=2$ and $k=3$ demonstrate cases where multiple solutions were found, indicating the presence of multiple basins of attraction even for small numbers of clusters. }
\end{figure*}

Beyond identifying local minima, we aim to assess the diversity of solutions and the prevalence of suboptimal local minima, providing deeper insights into the structure and multimodality of the landscape. To this end, we proceed as follows:
\begin{enumerate}
    \item Aggregate all final solutions obtained from the three local search algorithms.
    \item Perform hill-valley tests (with a number of 4 intermediary points) between all pairs of solutions to detect the presence of intermediate minima. 
    \item Construct a solution network, where nodes represent solutions and edges indicate the existence of an intermediate minimum between them.
    \item Identify fully connected subgraphs (cliques) within this network to group structurally similar solutions.
    \item Select representative solutions (those with the best objective value) from each clique.
\end{enumerate}

After identifying the representative solutions of each clique, we perform a pairwise comparison to investigate whether the associated clusters originate from the same basin of attraction. For each pair of representatives, we proceed as follows: starting from the representative with the worse objective value, we perform a simple local search by sequentially generating perturbed points in its vicinity. At each iteration, small random perturbations are applied, and the new point is accepted if it improves the objective value. This process continues until either (i) the search point moves sufficiently close to the better representative in the search space, suggesting that the two solutions belong to the same basin, or (ii) no further improvements can be found through perturbations, indicating that the solutions likely reside in different basins.
This procedure enables us to characterize not only the diversity of the local minima but also the connectivity between basins, offering deeper insight into the ruggedness and structure of the underlying search landscape.

We applied the previously described procedure to the final solutions obtained by the local search algorithms. As illustrated in Figure~\ref{fig:heatmap-diversity}, the results reveal a high degree of solution diversity across problem instances and dimensions. Some problems yield a single solution across runs, while others produce multiple distinct solutions, indicating the presence of several basins of attraction. These findings offer some evidence of multi-modality in the underlying landscapes.

Moreover, we observe that the number of basins increases with problem dimensionality, which matches with the intuition that higher-dimensional landscapes become more rugged and fragmented. This trend is also visible in Figures~\ref{fig:ruspini-k2} and \ref{fig:ruspini-k3}, where we present the solutions found on the Ruspini dataset (function F8) for $k=2$ and $k=3$ clusters, respectively. For $k=2$ (Figure~\ref{fig:ruspini-k2}), two distinct solutions are identified, while for $k=3$ (Figure~\ref{fig:ruspini-k3}), four unique solutions emerge.

\subsection{Breaking the Symmetry}

As a final perspective on clustering problems for optimization algorithms, we focus in more detail on their inherent symmetry. To illustrate these symmetries, we use a simple 1-dimensional dataset with $k=2$, to have a search-space with $2$ dimensions and $2$ regions of symmetry\cite{gallagher2016towards,hajari2024searching}. Since symmetry regions imply multi-modality, this could potentially hinder the performance of an optimization algorithm, especially when a population-based algorithm starts the search in multiple symmetrical regions, potentially leading to redundant search steps.

Because of this, it could be beneficial to find a representation of this problem which removes the symmetry. In essence, this would involve creating a (bijective) transformation function $t: [0,1]^m\rightarrow S\subset [0,1]^m$ where $S$ is a single region of symmetry. The problem of finding such a transformation has commonalities with the problem of sampling in a unit-simplex, for which the Dirichlet distribution can be used. As such, we can make use of a modified stick-breaking procedure:

We let $$t(x_1) = F^{-1}_{\text{Beta}}(x_{0} \mid 1, X)$$ and $$t(x_i) = t(x_{i-1}) + (1 - t(x_{i-1})) \cdot F^{-1}_{\text{Beta}}(u_{i} \mid 1, X - i)$$ for each $i \in \{2\dots k\}$. Here, $F^{-1}_{\text{Beta}}$ refers to the inverse CDF of the Beta-distribution with $\beta=X$. 

\begin{figure}
    \centering
    \includegraphics[width=0.95\linewidth]{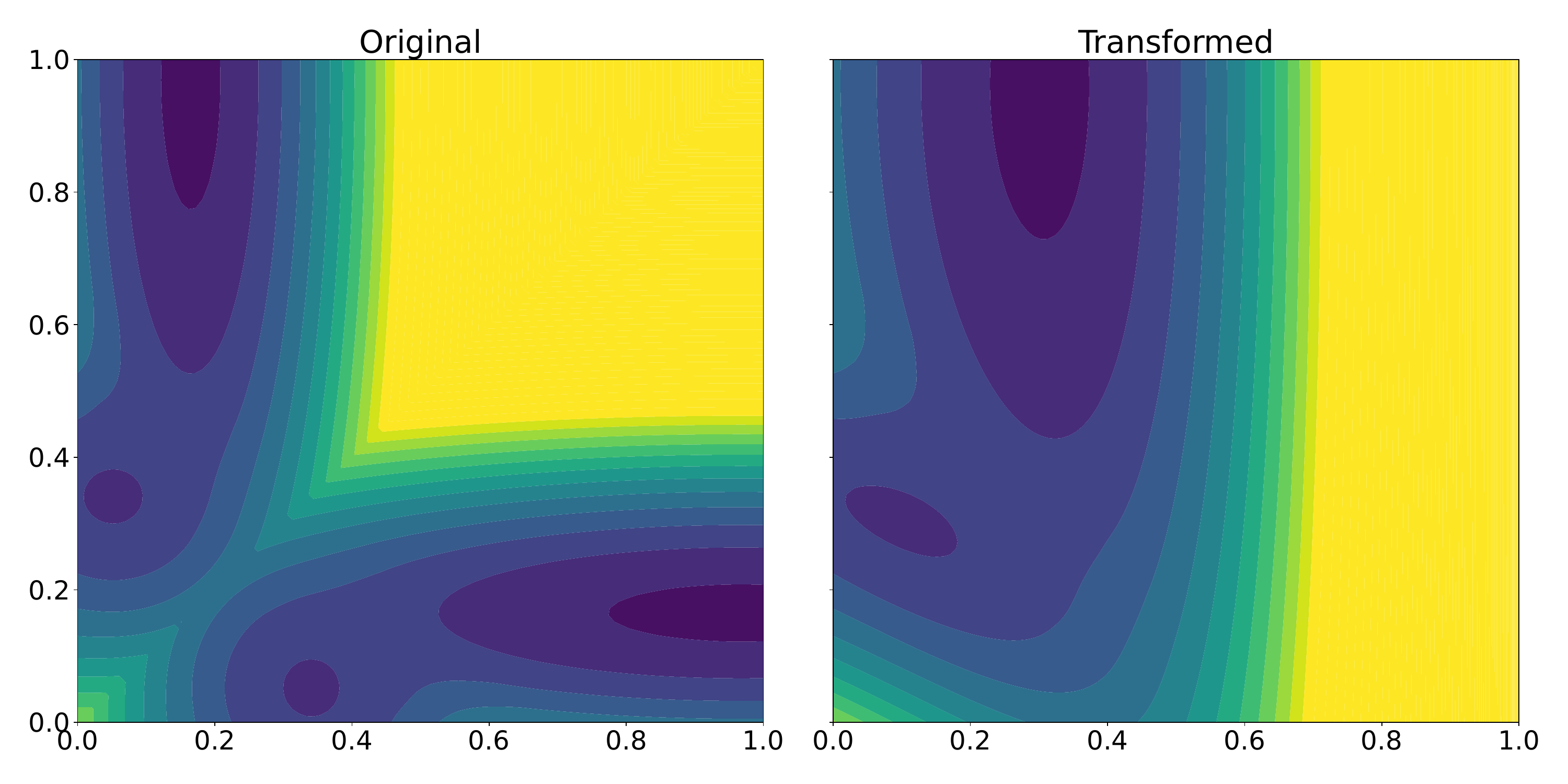}
    \caption{Example of Original and Transformed version of a 1-dimensional clustering problem with $k=2$.}
    \label{fig:transform_example}
\end{figure}

To illustrate this transformation on our 2-dimensional search space, we show both the original and transformed version of a clustering problem (with $d=1, m=k=2$) in Figure~\ref{fig:transform_example}. We note that to generalize this transformation to higher values of $d$, the transformation can be applied to the first component of each cluster center instead, while leaving the remaining coordinates intact. 

\begin{figure}
    \centering
    \includegraphics[width=0.95\linewidth]{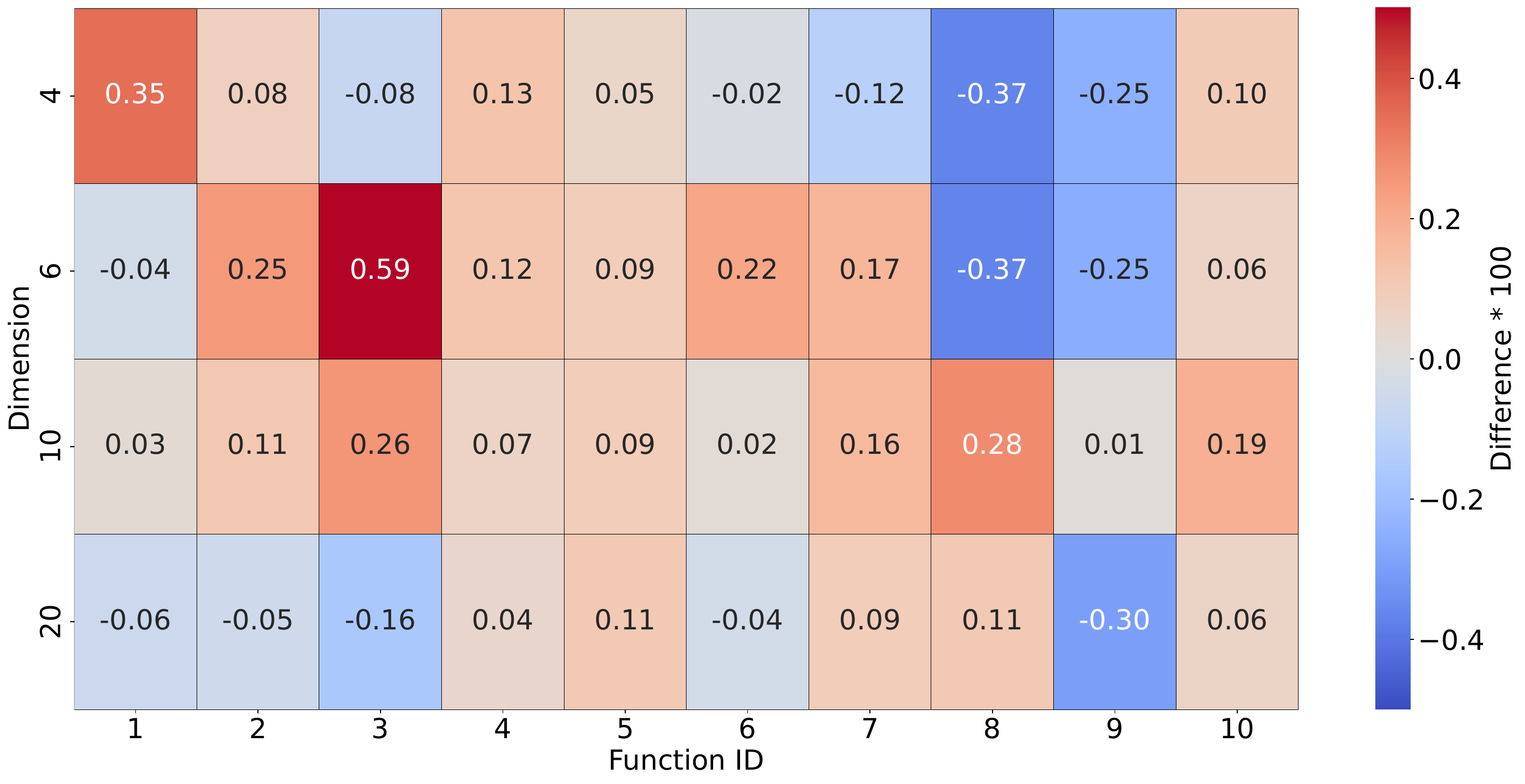}
    \caption{Differences in mean performance of \cma on the original vs transformed space. Negative values correspond to functions where the original representation led to better performance.}
    \label{fig:transform_perf}
\end{figure}

To verify whether this transformation has a positive impact on the performance of optimization algorithms, we benchmark the default configuration of \cma, both with and without this transformation, on each of the 40 problems in our proposed suite. We perform 25 repetitions on each problem, and compare the final function value found between the two setups. Their relative differences are visualized in Figure~\ref{fig:transform_perf}.

We can see from Figure~\ref{fig:transform_perf} that our proposed transformation results in similar performance to that of the default problem representation. However, there are some problems where the performance becomes worse, which might to suggest that these transformed search spaces become harder to navigate for this algorithm. When looking at the definition of the transformation, we should note that the first transformed variable depends only on $x_0$, but the second depends on both $x_1$ (for the Beta-distribution) and $t_0$ (for the weighting). Generally, each variable in the transformed space depends on the variables before it, inducing a hierarchy to the search variables. This hierarchical nature is inherently challenging for most population-based algorithms, so we can conclude that our proposed transformation does not necessarily result in an easier to optimize representation for the clustering problem. 

\section{Conclusions and Future Work}

In this paper, we proposed and analyzed a suite of optimization problems based on data clustering. These problems provide optimization challenges not commonly addressed in existing problem suites, but which are present in a variety of real-world domains. By performing a selection of benchmarking studies, we showcased the diversity of our problem set in relation to the commonly used BBOB suite, both from an algorithmic performance and landscape characteristic perspective. It is however still an open question how this compares to more practical optimization problems with similar symmetry characteristics in different real-world domains. 

With the introduction of this problem suite, and its integration with the IOHprofiler framework, we aim to make it easier for other researchers to utilise a wider variety of problems for algorithm benchmarking as well as promote further research on problems with neutrality and permutation symmetries. While we observe some potential benefits to symmetry-breaking transformations, further systematic research is required to determine the extent of their usefulness. 

Since more clustering problem instances can easily be specified with additional datasets and scaled to higher dimensionalites by increasing $k$, we, in addition to the problem suite, also provide a simple problem generator. Since this generator comes with increased flexibility in terms of the clustering setup, it would be especially interesting to analyze how changes to the setup, e.g. changing the error measure to a maximum error per cluster center, would impact the fundamental landscape properties of the resulting optimization problem, and in turn how this impacts the relative performance of different types of optimization algorithms. This could help fine-tune algorithms to address challenges in specific problem domains.

\begin{acks}
    Parts of this work were performed while Diederick Vermetten was employed at LIACS, Leiden University. 
This project was partially supported by the European Union (ERC, ``dynaBBO'', grant no.~101125586). Views and opinions expressed are however those of the author(s) only and do not necessarily reflect those of the European Union or the European Research Council Executive Agency. Neither the European Union nor the granting authority can be held responsible for them.
\end{acks}

\bibliographystyle{ACM-Reference-Format}

\bibliography{bibliography}

\end{document}